\documentclass[conference]{IEEEtran}
\IEEEoverridecommandlockouts
\usepackage{graphics} 
\usepackage{epsfig} 
\usepackage{mathptmx} 
\usepackage{times} 
\usepackage{amsmath} 
\usepackage{amssymb}  
\usepackage{cite}
\usepackage{amsmath,amssymb,amsfonts}
\usepackage{algorithmic}
\usepackage{graphicx}
\usepackage{textcomp}
\usepackage{xcolor}
\usepackage{lipsum}
\usepackage{array}
\usepackage{booktabs}
\usepackage{tabularx}
\usepackage{multicol}
\usepackage{multirow}
\usepackage{xcolor}
\usepackage{amssymb}
\usepackage{amsmath}
\usepackage{stfloats}
\usepackage{mathdots}
\usepackage{graphicx} 
\usepackage{caption}
\usepackage{subfigure}
\usepackage{algorithm}
\usepackage{bm}
\usepackage{listings}

\usepackage{xcolor}

\definecolor{codegreen}{rgb}{0,0.6,0}
\definecolor{codegray}{rgb}{0.5,0.5,0.5}
\definecolor{codepurple}{rgb}{0.58,0,0.82}
\definecolor{backcolour}{rgb}{0.95,0.95,0.92}

\lstdefinestyle{mystyle}{
    backgroundcolor=\color{backcolour},   
    commentstyle=\color{codegreen},
    keywordstyle=\color{magenta},
    numberstyle=\tiny\color{codegray},
    stringstyle=\color{codepurple},
    basicstyle=\ttfamily\footnotesize,
    breakatwhitespace=false,         
    breaklines=true,                 
    captionpos=b,                    
    keepspaces=true,                 
    numbers=left,                    
    numbersep=5pt,                  
    showspaces=false,                
    showstringspaces=false,
    showtabs=false,                  
    tabsize=2
}

\begin{document}
\title{Attacking Digital Twins of Robotic Systems to Compromise Security and Safety\thanks{This paper was supported by the Manchester Metropolitan University.}
}

\author{Christopher Carr$^{1}$, Shenglin Wang$^{2}$, Peng Wang$^{*,1}$  and Liangxiu Han$^{1}$
\thanks{*Correspondence to Peng Wang: p.wang@mmu.ac.uk}
\thanks{$^{1}$Christopher Carr, Peng Wang, and Liangxiu Han are with the Department of Mathematics and Computing,
        Manchester Metropolitan University, M15 6BH, United Kingdom
        {\tt\small \{p.wang, l.han\}@mmu.ac.uk}, christopher.carr@stu.mmu.ac.uk}%
\thanks{$^{2}$Shenglin Wang is with the Department of Automatic Control and Systems Engineering, The University of Sheffield,
        Sheffield, S10 2TN, UK
        {\tt\small swang119@sheffield.ac.uk}
        }
}

\maketitle

\begin{abstract}
Security and safety are of paramount importance to human-robot interaction, either for autonomous robots or human-robot collaborative manufacturing. The intertwined relationship of security and safety has imposed new challenges on the emerging digital twin systems of various types of robots. To be specific, the attack of either the cyber-physical system or the digital-twin system could cause severe consequences to the other. Particularly, the attack of a digital-twin system that is synchronised with a cyber-physical system could cause lateral damage to humans and other surrounding facilities. This paper demonstrates that for Robot Operating System (ROS) driven systems, attacks such as the person in the middle attack of the digital-twin system could eventually lead to a collapse of the cyber-physical system, whether it is an industrial robot or an autonomous mobile robot, causing unexpected consequences. We also discuss potential solutions to alleviate such attacks.
\end{abstract}



\section{INTRODUCTION}

The concept of `digital twin' was first coined by John Vickers around 2004 and it originally referred to a virtual representation of a product in a digital form~\cite{gehrmann2019digital}. The content of digital twin has been massively enriched during the past decade, and nowadays, it mostly refers to the counterpart `Digital-Twin System (DTS)' of a Cyber-Physical System (CPS). In a current DTS-CPS dual system, a fine granular module of the CPS can find its digital counterpart in the DTS. These modules are connected through network infrastructures, to enable data transfer and communication. After years of development, DTS is capable of incorporating data collected from sensors installed on CPS to simulate the process on CPS, and then forecast behaviours of the CPS, to help optimise the process, monitor the lifecycle of the CPS, and avoid incidents such as injuries to humans. NASA~\cite{glaessgen2012digital} and Siemens~\cite{siemens2021cobot}, etc. have been applying DTS in space exploration and complex system design, etc. 


Recently, digital-technique driven human-robot interaction has attracted broad attention in the context of autonomous driving~\cite{almeaibed2021digital} and manufacturing~\cite{peng2021cobot,gehrmann2019digital}, where safety and security is critical as robots and humans need to work physically in shared spaces. In such cases, the deployment of a DTS would help to shift direct human-robot contacts in CPS to the interaction between humans and a DTS, where injuries are unlikely to occur. Such safety related topic of DTS has been widely investigated during the past few years. For instance, our previous work exploited deep learning techniques to detect and separate humans and robots in a human-robot collaborative (HRC) manufacturing scenario, with beliefs on separation results provided along to counter model and measurement uncertainties to further ensure safety ~\cite{peng2021cobot}. Malik and Brem ~\cite{malik2021digital} propose to identify and resolve potential safety risks of solutions in DTS before being deployed onto CPS, which helps to achieve significant safety performance improvement with a massive reduction of physical experiments. This also opens door to teleoperation of CPS working at a distance or in hazardous environments ~\cite{luckcuck2021principles}, to ensure safety of human operators. Recently, DTS has been applied in CPS reconfiguration with the increasing demand for customised manufacturing, where a CPS needs to be frequently reconfigured for individualised product manufacturing ~\cite{leng2020digital}. 

\begin{figure*}[ht]
    \centering
    \includegraphics[width=0.75\linewidth]{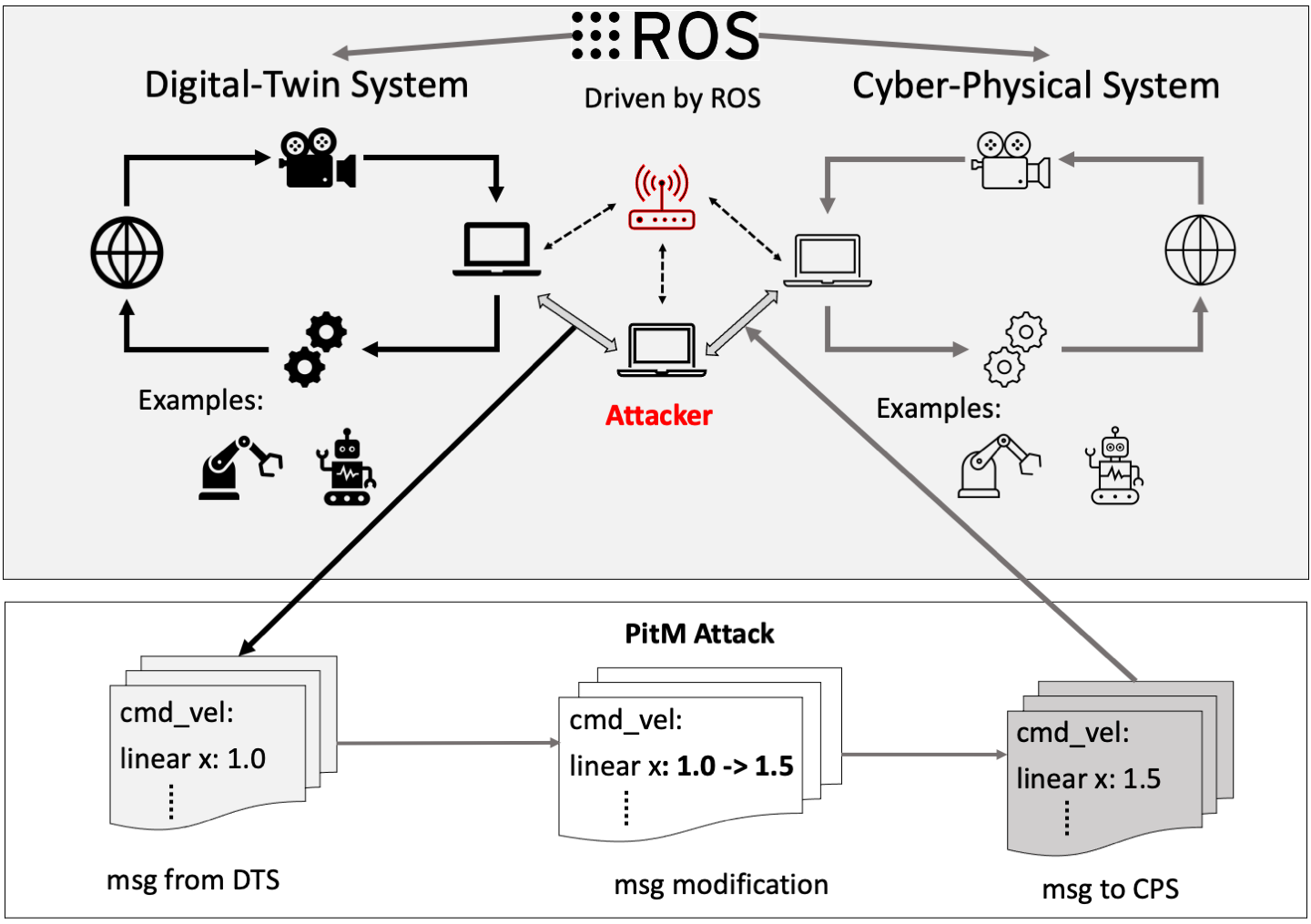}
    \caption{Illustration of attacks on the information flow from the DTS to the CPS}
    \label{fig:attack}
\end{figure*}

However, no solutions are provided to deal with safety issues induced by security in the above papers. Suhail et al.~\cite{suhail2022security} investigated security attacks and solutions for digital twin systems. A multitude of attack cases and solutions were proposed. Hearn and Rix~\cite{hearn2019cybersecurity} raised the concerns that while DTS is promising, it also increases the attack surface of CPS as hackers can now attack CPS via DTS. Overall, the cloud-based or remote-server based implementation of DTS has brought in new challenges~\cite{susila2020impact}. Nevertheless, these papers discuss security and safety of DTS-CPS system in general, without specific emphasis on DTS of robotic CPS.
Currently, a lot of robots are driven by the Robot Operating System (ROS), where no default security measures are provided to counter attacks. In addition, the ROS mechanism of allowing any subscriber nodes (computers) within the same Local Area Network (LAN) to access published topics makes such robotic CPS prone to Person in the Middle (PitM) attacks ~\cite{teixeira2020security}. Particularly, when the information flow from the DTS to the CPS is attacked, as shown in Figure \ref{fig:attack}, the robot will behave unexpectedly, conceivably causing injuries to humans. There are works about using DTS of robots to deal with safety issues induced by security~\cite{almeaibed2021digital,suhail2022security}, we will discuss them further in related work! 


This paper will focus on how a ROS driven robotic CPS can be attacked through its DTS. We raise two examples where the information flow from the DTS to the CPS is attacked, leading to potential safety issues. In the first example, the CPS is a TurtleBot 3 robot and the DTS is implemented in Gazebo. In the second example, the CPS is a Universal Robot 10, and the DTS is implemented in Unreal Engine. Our main contributions include 1) DTS-CPS attack cases are raised for both mobile and industrial robots. 2) PitM attacks of two robotic systems and their digital twin systems are carried out, demonstrating the vulnerabilities of ROS based robotic DTS-CPS systems. Solutions to deal with such attacks will appear in our furture works!

The remainder of the paper is organised as follows. Some related works are introduced in Section II. Section III elaborates on the attack case studies. Discussions and potential solutions are provided in Section IV, and the paper is concluded in Section V.

\section{RELATED WORK}

ROS was not designed with security in mind in the first place. However, it has become the \textit{de facto} operating system for robots and the growing deployment of ROS in industry and other safety-critical sectors such as caring constantly raises the question of `can ROS be used securely in real-life applications?'. This eventually led to the development of ROS2, where security has been taken into consideration from design, but it will take years before ROS2 prevails over ROS ~\cite{mayoral2020can}. Therefore, there is still the need to look into the security aspect of ROS.

The vulnerability, as well as the robustness of ROS when it is under attack, has been investigated shortly after the creation of ROS in 2007~\cite{mcclean2013preliminary}. Vctor et al.~\cite{mayoral2020can} have thoroughly and systematically investigated the security of ROS, and categorised possible attacks that compromise ROS security into four classes:
\begin{itemize}
    \item \textit{Attack 1} $(A_1)$: Targeting ROS-Industrial and ROS core packages.
    \item \textit{Attack 2} $(A_2)$: Disrupting ROS-Industrial communi- cations by attacking underlying network protocols.
    \item \textit{Attack 3} $(A_3)$: Person-In-The-Middle (PitM) attack on a ROS control station.
    \item \textit{Attack 4} $(A_4)$: Exploiting known vulnerabilities in a robot endpoint to compromise the ROS network.
\end{itemize}
\noindent
The authors designed the above four attacks to compromise ROS computational graph and demonstrated that ROS is prone to all four types of attacks. The authors concluded that `\textit{with the current status of ROS, it is hardly possible to guarantee security without additional measures}', which further confirms that we should be cautious in applying ROS in safety-critical scenarios. However, the authors mainly focus on the vulnerability of ROS rather than a DTS.

In a DTS-CPS scenario, communication between the two systems introduces new security challenges. One example is the attack of information flow from DTS to CPS as shown in figure  \ref{fig:attack}. This is raised because humans directly interact with CPS, and any malicious modification of the information flow from DTS to CPS will cause unexpected and dangerous behaviours of the robot - some of which can lead to injuries to humans. Architectures have been designed to resolve the issue, for instance, Christian and Martin~\cite{gehrmann2019digital} developed an adversarial model for DTS security ensurance for industrial automation and control systems. They have demonstrated that the adversarial model is promising for protecting the CPS while opening them up for external data sharing and access. The adversarial model was successfully tested on `\textit{a small but real production case}'. The model needs further work for a fully working system including intrusion detection, access control, security analysis, etc. Sadeq et al. have discussed how DTS could help to promote safety and security in autonomous vehicles~\cite{almeaibed2021digital}.

This work stems from \cite{gehrmann2019digital,mayoral2020can, almeaibed2021digital}, and focuses on analysing the vulnerability of ROS based DTS-CPS systems when facing PitM attacks. Two case studies include a mobile robot with its digital system and an industrial robot with its digital system are introduced as follows.

\section{CASE STUDIES}

\subsection{PitM Attack of an Autonomous Robot via Its Digital Twin System}
Velocity control is one of the most important factors for ensuring safety of autonomous robots. Ideally, a robot should either decide by itself or be controlled by humans (via DTS) to change its linear and angular velocities to safely co-exist in human society. 

For a ROS driven DTS-CPS dual system, one potential \textbf{scenario} is that a robot in DTS is moving along the $x$-axis while communicating its velocity to the robot in the CPS, such that the real robot synchronously moves along the $x$-axis as well. The information flow from DTS to CPS could be attacked by a PitM approach. To be specific, \textbf{Listing 1} shows the \textit{Twist} message used in ROS for controlling the linear and angular velocities of a robot. When the digital robot is moving along $x$-axis with velocity 1.0 m/s, the \textit{Twist} message will be like \textbf{Listing 2}, where one can see that the $x$ component of the linear velocity is modified to 1.0 m/s.

\begin{lstlisting}[language=Python, caption=Data structure of the \textit{Twist} message in ROS]
geometry_msgs/Vector3 linear 
  float64 x #x-axis velocity 
  float64 y
  float64 z
geometry_msgs/Vector3 angular
  float64 x
  float64 y
\end{lstlisting}

\begin{lstlisting}[language=Python, caption=\textit{Twist} message of the digital robot moving along $x$-axis]
geometry_msgs/Vector3 linear 
  float64 x: 1.0 #x-velocity in DTS
  float64 z: 0.0
geometry_msgs/Vector3 angular
  float64 x: 0.0
  float64 y: 0.0
  float64 z: 0.0
\end{lstlisting}

When the digital robot communicates its velocity information to the real robot, the information flow could be intercepted and maliciously modified. For instance, the velocity along the $x$-axis could be modified to 1.5 m/s, and the message received by the real robot will be like \textbf{Listing 3}. This will cause issues if the real robot was expected to move with a 1.0 m/s velocity along the $x$-axis to ensure safety.

\begin{lstlisting}[language=Python, caption=Modified \textit{Twist} message received by the real robot]
geometry_msgs/Vector3 linear 
  float64 x: 1.5 #x-velocity in CPS
  float64 y: 0.0
  float64 z: 0.0
geometry_msgs/Vector3 angular
  float64 x: 0.0
  float64 y: 0.0
  float64 z: 0.0
\end{lstlisting}

Implementation details of the PitM attack is given in Algorithm~\ref{alg:AR-Attack}. The \textit{Twist} message is published on the \textit{cmd\_vel} ROS topic by the digital robot and the real robot subscribes to the \textit{cmd\_vel} topic. We also assume that the DTS-CPS dual system works under the same LAN. We denote the \textit{Twist} message published by the DTS as dts\_msg and the message received by the CPS as cps\_msg. For each step, multiple tools could be used, we have listed some for reference.
\begin{algorithm}[tbh]
	\caption{PitM Attack of a DTS-CPS Dual System}
	\label{alg:AR-Attack}
	\begin{algorithmic}[1]
		\renewcommand{\algorithmicrequire}{\textbf{Input:}}
		\REQUIRE dts\_msg
		\renewcommand{\algorithmicrequire}{\textbf{Output:}}
		\REQUIRE cps\_msg
		\STATE Capture IP and MAC addresses of the targeted machine and ROS master machine. - \textit{nmap, netdiscover}
		\STATE Perform ARP attacks on both machines, to assume the IP of the other (intercept the packets). - \textit{arpspoof, ettercap, bettercap}
		\STATE Filter intercepted packets according to IP addresses, message transfer protocols including TCP and UDP, and ROS topic flags. - \textit{wireshark, scapy}
		\STATE Deciphering the found dts\_msg ROS messages and modify it to generate cps\_msg - \textit{scapy}
		\STATE Forward all packets, including the modified cps\_msg to both machines. - \textit{scapy}

	\end{algorithmic}
\end{algorithm}

\subsection{PitM Attack of an Industrial Robot via Its Digital Twin System}

For industrial robots, we have chosen a human-robot collaboration case where safety of human operators is critical. The chosen ROS topic to intercept and modify is shown in \textbf{Listing 4}. Note that this message has a very complex data structure. We therefore use indentation to indicate the hierarchy between data structures. For instance, the \textit{points} of type `trajectory\_msgs/JointTrajectoryPoint[]' is comprised of 

\begin{lstlisting}[language=Python]
    float64[] positions;
    float64[] velocities;
    float64[] accelerations;
    duration time_from_start.
\end{lstlisting}
\textit{positions} contains radian position information of the six joints of a 6DoF Universal Robot 10 robot. The joints include one elbow joint, denoted as `elbow\_joint' in \textit{positions}, two shoulder joints including `shoulder\_lift\_joint' and `shoulder\_pan\_joint', and three wrist joints, i.e., `wrist\_1\_joint', `wrist\_2\_joint', and `wrist\_3\_joint'. For initialisation, the \textit{velocities} and \textit{accelerations} are set to \textbf{0} for both DTS and CPS, and the DTS gets its \textit{positions} from the CPS. When DTS gains its control over CPS, the DTS will change `shoulder\_pan\_joint' such that both the digital robot and the real robot would rotate their shoulder for a random angle.

Algorithm \ref{alg:AR-Attack} is used to intercept and modify the information flow from the DTS to the CPS again. To be specific, the \textit{positions} information sent from DTS is intercepted and the rotation angle of `shoulder\_pan\_joint' is modified, such that the real robot keeps rotating in one direction regardless of the movement of the digital robot. If a human operator is safely collaboration with the robot in the route defined by the message from DTS, then the route with wrong direction will conceivably cause injuries to the human operator. Furthermore, it also brings potential damages to surrounding facilities.

\begin{lstlisting}[language=Python, caption=control\_msgs/FollowJointTrajectoryActionGoal.msg Structure]
std_msgs/Header header
actionlib_msgs/GoalID goal_id
control_msgs/FollowJointTrajectoryGoal goal
    control_msgs/JointTolerance[] path_tolerance
    control_msgs/JointTolerance[] goal_tolerance
    duration goal_time_tolerance
    trajectory_msgs/JointTrajectory trajectory
        std_msgs/Header header
        string[] joint_names
        trajectory_msgs/JointTrajectoryPoint[] points
            float64[] positions #Data being attacked!
            float64[] velocities
            float64[] accelerations
            duration time_from_start
    \caption{}
    \label{lst:hrc}
\end{lstlisting}

\section{DISCUSSIONS}
The two case studies have shown that ROS based DTS-CPS dual systems are prone to malicious attacks such as PitM in our cases. The information flow from DTS to CPS can be intercepted and modified to cause intentional injuries to targets in CPS. There are purposely designed digital systems to avoid such attacks, but they are only applicable in small cases~\cite{gehrmann2019digital}. Other possible solutions toward safe and secure DTS-CPS dual systems include encrypting information flow between the two systems, or gradually migrant to systems such as ROS2 to get better security protection in safety-critical cases!

\section{CONCLUSIONS}

Case studies regarding the security of two robotic DTS-CPS dual systems are investigated and discussed. Despite the increasing demand for digital twin systems in manufacturing, etc., their security and the safety issues induced by security remain as challenges, particularly for safety-critical scenarios. We demonstrated through the two case studies that information flow from the DTS to the CPS (and vice versa) can be easily intercepted and modified with the PitM attack, leading to unexpected behaviors of the dual systems. This alerts the wide ROS based DTS community that safety and security should be seriously dealt with, particularly when the application is safety-critical.

\bibliographystyle{IEEEtran}
\bibliography{AttackDigitalTwins}

\end{document}